\documentclass[lettersize,journal]{IEEEtran}
\usepackage{amsmath,amsfonts}

\usepackage{algorithmic}
\usepackage{algorithm}
\usepackage{array}
\usepackage[caption=false,font=normalsize,labelfont=sf,textfont=sf]{subfig}
\usepackage{textcomp}
\usepackage{stfloats}
\usepackage{url}
\usepackage{verbatim}
\usepackage{graphicx}
\usepackage{cite}
\usepackage[colorinlistoftodos]{todonotes}
\usepackage{booktabs}
\usepackage{threeparttable}
\usepackage{graphicx}
\usepackage{xcolor}
\usepackage{tikz}
\usetikzlibrary{arrows.meta, positioning}

\tikzset{
  layer/.style={rectangle, draw=black, rounded corners, thick, minimum width=5.5cm, minimum height=1.2cm, align=center},
  arrow/.style={-{Latex}, thick},
  node distance=1.5cm
}

\begin{document}

\title{Fair and Safe: A Real-Time Hierarchical Control Framework for Intersections}

\author{Lei Shi, Yongju Kim, Xinzhi Zhong, Wissam Kontar, Qichao Liu, Soyoung Ahn
}

\markboth{Journal of \LaTeX\ Class Files,~Vol.~14, No.~8, August~2025}%
{Shell \MakeLowercase{\textit{et al.}}: A Sample Article Using IEEEtran.cls for IEEE Journals}


\maketitle

\begin{abstract}
Ensuring fairness in the coordination of connected and automated vehicles at intersections is essential for equitable access, social acceptance, and long-term system efficiency, yet it remains underexplored in safety-critical, real-time traffic control. This paper proposes a fairness-aware hierarchical control framework that explicitly integrates inequity aversion into intersection management. At the top layer, a centralized allocation module assigns control authority (i.e., selects a single vehicle to execute its trajectory) by maximizing a utility that accounts for waiting time, urgency, control history, and velocity deviation. At the bottom layer, the authorized vehicle executes a precomputed trajectory using a Linear Quadratic Regulator (LQR) and applies a high-order Control Barrier Function (HOCBF)-based safety filter for real-time collision avoidance. Simulation results across varying traffic demands and demand distributions demonstrate that the proposed framework achieves near-perfect fairness, eliminates collisions, reduces average delay, and maintains real-time feasibility. These results highlight that fairness can be systematically incorporated without sacrificing safety or performance, enabling scalable and equitable coordination for future autonomous traffic systems.
\end{abstract}


\section{Introduction}


Fairness is an increasingly critical aspect of modern transportation systems \cite{pereira2017distributive, beyazit2011evaluating, yin2022maximizing}, particularly with the emergence of connected and automated vehicles (CAVs)\cite{martinez2022transport}. In conventional traffic environments, fairness—defined as the equitable allocation of road resources\cite{wu2024dynamic}—is often implicitly managed through social norms, established traffic rules\cite{wu2017delay}, and informal human interactions. In contrast, automated transportation systems rely on algorithmic and computational decision-making, requiring explicit definitions and enforcement mechanisms for fairness. Failure to do so can result in longer delays for specific user groups, reduced public acceptance, and persistent biases in system decisions and outcomes \cite{ye2022fairlight, du2024felight}. Ensuring fairness in traffic coordination thus not only enhances user experience but also aligns transportation operations with broader ethical principles and social justice objectives.

However, pursuing fairness inherently introduces challenges, particularly in relation to safety \cite{li2022trade}. Equitable allocation of road resources may conflict with optimal safety protocols \cite{7136395}. For instance, algorithms that aim to distribute delays evenly across vehicles could inadvertently compromise essential safety margins such as minimal distances or reaction times. Balancing fairness and safety therefore requires careful evaluation of potential trade-offs and system-level impacts.

These challenges are further amplified with the emergence of CAVs \cite{wang2021review}. As CAVs make high-frequency (i.e., millisecond-level), networked decisions, one vehicle's action can rapidly influence others, causing biases or errors to propagate through traffic system. Even minor inaccuracies in perception or control failure can amplify into major system-wide unfairness or safety risks. Ensuring fairness and safety under such tightly coupled dynamics is therefore essential to the sustainable deployment and public acceptance of CAV technologies.


To address these challenges, multi-vehicle coordination strategies are required. As a typical multi-agent system, CAV control can be organized in centralized and/or decentralized frameworks \cite{8352646}. Centralized control allows a central authority to coordinate all vehicles and enforce fairness uniformly. However, it faces scalability limitations, slower computation, and vulnerability to single-point failures that hinder real-time safety-critical operations. In contrast, decentralized control distributes decision-making across individual vehicles, enhancing scalability and robustness to failures \cite{mahbub2020decentralized,4177054}. Yet it also brings new challenges—ensuring fair resource allocation, maintaining safety in dense traffic, achieving efficiency, and preserving near-optimal performance without global coordination—particularly at intersections.

Intersections represent one of the most complex environments for balancing fairness and safety. Conflicting vehicle trajectories, limited space, and competing intentions make real-time coordination challenging. In such scenarios, aggressive maneuvers or failure to yield can not only heighten collision risk but also lead to inequitable delays among vehicles. These problems can become even more pronounced at intersections with unbalanced traffic demand, where vehicles from certain approaches may experience disproportionately longer delays. 
Such conditions underscore the necessity of developing intelligent coordination strategies for CAVs that can jointly ensure fairness and safety at intersections \cite{9837467, 10159571}.

\begin{figure}[htbp]
\captionsetup{font={footnotesize}}
\centering
\includegraphics[width=1.0\linewidth]{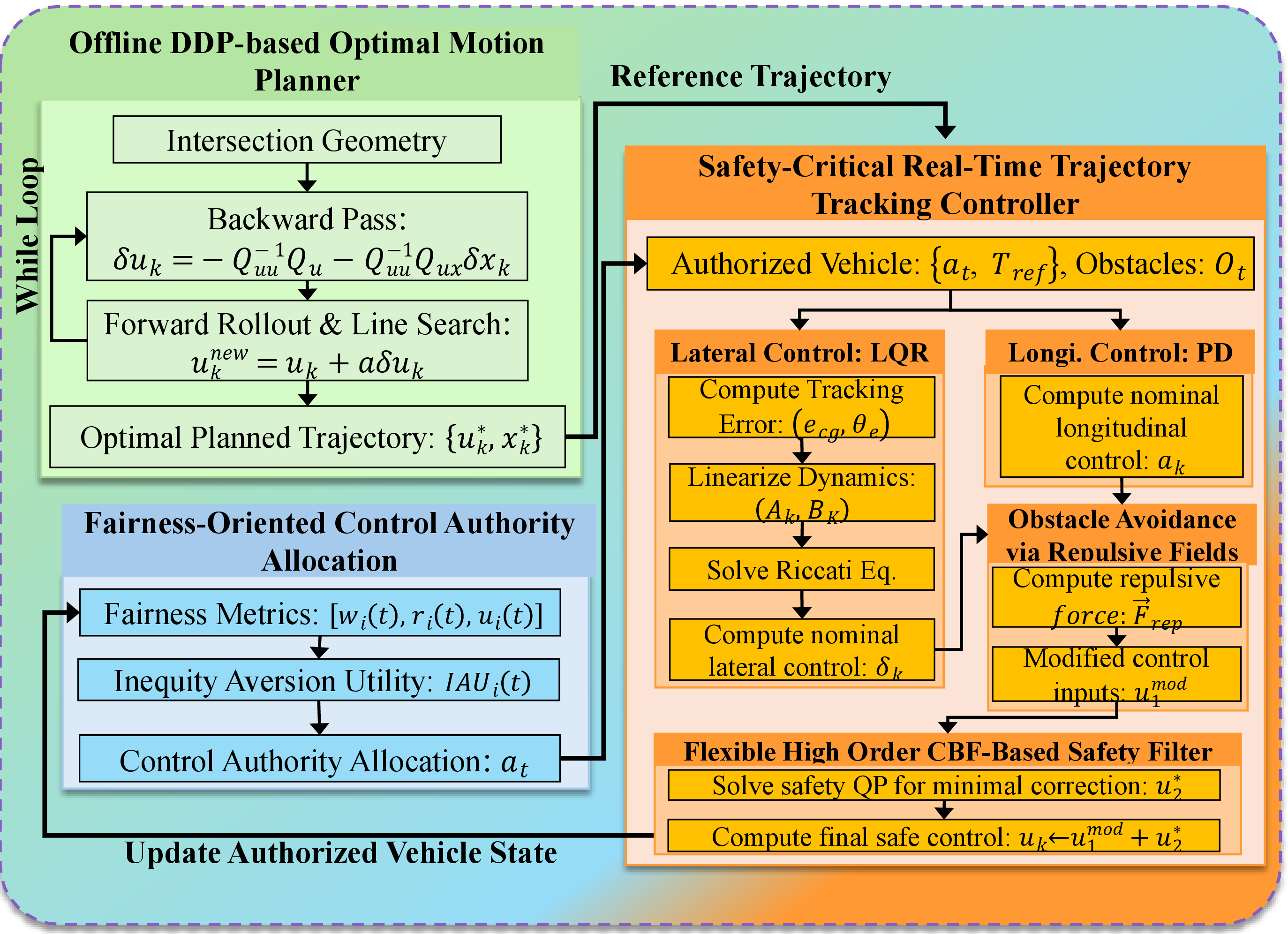}
\label{fig:control_framework}
\vspace{-0.6cm}
\caption{Overview of the proposed hierarchical control framework. The system first generates reference trajectories using offline DDP-based optimal motion planner. Then, a fairness-oriented control authority allocation module—driven by an inequity aversion utility—rapidly determines the authorized vehicle, enabling hybrid centralized planning and decentralized execution. Finally, the real-time trajectory tracking controller executes the planned trajectory while ensuring safety through LQR/PD control, repulsive fields, and a flexible high-order CBF-based safety filter.}
\end{figure}

In this paper, we propose a novel hierarchical control framework for CAVs at intersections that explicitly balances fairness and safety while ensuring computational efficiency and scalability. The framework consists of two tightly integrated layers with distinct but complementary roles, as illustrated in Fig. \ref{fig:control_framework}:

\begin{itemize}
\item \textbf{Top Layer – Fairness-Oriented Control Authority Allocation (Centralized):} A centralized decision-making module assigns \emph{control authority} to a single vehicle at each time step. Here, \emph{control authority} refers to the exclusive right of a vehicle to follow its planned trajectory while considering all other vehicles as dynamic obstacles, i.e., vehicles without current authority cannot execute their own planned paths. This module evaluates a utility function incorporating multiple fairness metrics inspired by inequity aversion, thereby ensuring equitable treatment across all vehicles over time.

\item \textbf{Bottom Layer – Safety-Critical Real-Time Trajectory Tracking with Optimal Planning (Decentralized):} Once granted control authority, each vehicle executes its locally planned trajectory. The trajectory is precomputed offline using Differential Dynamic Programming (DDP) and tracked in real time using a discrete-time Linear Quadratic Regulator (LQR). To ensure safety under dynamic and uncertain conditions, Control Barrier Functions (CBFs) are incorporated as a safety filter within the controller.
\end{itemize}

This hierarchical design leverages centralized coordination in the top layer to ensure global fairness and decentralized execution in the bottom layer to achieve safety, real-time responsiveness, and scalability. Its modular structure enables independent development, formal verification, and practical deployment in embedded systems.

The main contributions of this work are summarized as follows:
\begin{enumerate}
\item We design a computationally efficient, fairness-oriented control authority allocation mechanism based on inequity aversion utility. The centralized mechanism rapidly assigns control authority while delegating control input generation to the decentralized module. This division of roles ensures both fairness and safety by combining global coordination with responsive, robust decentralized execution, and guarantees real-time feasibility. 
\item We develop a high-order and flexible CBF formulation that ensures collision avoidance under dynamic and uncertain traffic conditions.
\item We propose an integrated motion planning and control strategy that combines offline DDP-based trajectory generation with a real-time curvature-compensated LQR controller enhanced by a safety filter, achieving safe, robust, near-optimal, and computationally efficient trajectory tracking.
\item We validate the proposed framework through comprehensive simulations under various traffic demands and demand distributions, demonstrating its effectiveness in achieving fairness, safety, and efficiency at intersections.
\end{enumerate}


The remainder of this paper is organized as follows. Section \ref{sec:relatedwork} reviews related work. Sections \ref{sec:toplayer}-\ref{sec:bottomlayer} detail the proposed hierarchical control framework with fairness-oriented authority allocation, safety-critical trajectory tracking, and optimal motion planning. Section \ref{sec:experiments} reports experimental results, and Section \ref{sec:conclusion} concludes the paper.

\section{Related Work} \label{sec:relatedwork}

Our work integrates three key components of: (i) \emph{fairness-oriented resource allocation}, (ii) \emph{safety-critical control}, and (iii) \emph{offline trajectory planning}. In this section, we review relevant literature in each area and highlight the gaps our hierarchical framework is designed to address.

\subsection{Fairness‑Oriented Resource Allocation}
Early schemes adopt first–come–first–served (FCFS) policies and implicitly assume fairness in vehicle coordination \cite{liu2024gasoline}. FCFS is also frequently used in autonomous intersection management \cite{9217470, dresner2008multiagent}, often as a proxy for fairness. Subsequent work framed the crossing order as a non‑cooperative game, solving for Nash equilibria \cite{9810195, 10738130} that improve network throughput while respecting driver incentives.  More recent game‑theoretic approaches incorporate Jain’s index into the utility, leading to \emph{Fairness‑Aware Game Theoretic} (FGT) rules \cite{9458933,9447223}. However, most algorithms allocate resources through iterative best–response updates, which incur significant latency and offer no hard safety guarantees. They also couple fairness logic with low‑level control, limiting scalability. Our top‑layer allocator retains equity by maximizing an \emph{inequity‑aversion} utility in a single step, thereby sidestepping convergence issues while exposing a clean interface to downstream safety controllers.

\subsection{Safety-Critical Control via Control Barrier Functions}
Safety-critical control aims to ensure the satisfaction of safety constraints, but traditional penalty-based methods often lack formal guarantees and may fail under aggressive maneuvers. CBFs provide a Lyapunov–like certificate for forward invariance of safety sets \cite{8796030, 7782377}. High-Order CBFs (HOCBFs)\cite{9516971} extend the framework to higher relative-degree constraints such as inter-vehicle distance in bicycle models. Extensions for moving or deformable obstacles include adaptive CBFs \cite{9410332} and uncertainty-aware filters that blend reinforcement learning with safe control \cite{marvi2021safe}. Unlike static or slow-moving obstacles, vehicles are fast-moving agents with dominant longitudinal motion patterns. Time-varying CBFs remain relatively underexplored, and real-time formulations that explicitly capture velocity-dependent behaviors—such as those of fast-moving, predominantly longitudinal vehicles—are even more scarce.

\subsection{Offline Trajectory Planning}
Offline trajectory planning seeks to pre-compute dynamically feasible motion plans that downstream feedback controllers can track with minimal online burden.  Typical approaches include sampling-based search \cite{5980479}, direct collocation with nonlinear programming \cite{geiger2006optimal}, and gradient-based functional optimization such as CHOMP or STOMP.  While effective at producing collision-free paths, these approaches typically optimize over simplified kinematic models or coarse time discretizations, making the resulting motions less dynamic or optimal for a real vehicle. DDP\cite{6907001, 10160817} instead performs a second-order expansion of the nonlinear optimal-control problem, iteratively improving both the control sequence and a time-varying feedback gain. DDP provides fidelity by incorporating full nonlinear vehicle models, ensuring the generated trajectories respect physical constraints. Additionally, it achieves optimality by minimizing a user-defined cost function (e.g., energy, time, comfort), often outperforming sampling-based or direct-collocation methods.

Unlike prior works that have treated fairness, safety, and optimal motion planning as separate problems, our hierarchical framework leverages a newly proposed rapid control authority allocation method to synergistically integrate these components within a unified architecture. Meanwhile, the newly proposed Flexible HOCBFs rigorously ensure safety.

\section{Top Layer: Fairness-Oriented Control Authority Allocation} \label{sec:toplayer}

In this section, we introduce a novel fairness-driven approach for dynamically allocating control authority among vehicles at intersections. Unlike conventional traffic management systems that rely on fixed rules or simple heuristics, our method assigns control authority based on multiple fairness metrics.

\subsubsection{Problem Formulation}

Consider an intersection with a set of vehicles $\mathcal{V} = \{1, 2, \ldots, N\}$ approaching from various directions. At each discrete time step $t \in \{0, 1, \ldots, T-1\}$, the algorithm must decide which vehicle receives control authority. This decision directly influences each vehicle's ability to pass through the intersection safely and efficiently.

Let \( A_t \in \mathcal{V} \) denote the vehicle granted control authority at time \( t \). Our goal is to find an allocation policy \( \pi: t \mapsto A_t \) that maximizes fairness across all vehicles. The key challenge lies in defining fairness and identifying factors that should be considered. To this end, we utilize a fairness-aware utility—\textit{Inequity Aversion Utility (IAU)}—that integrates diverse fairness metrics and explicitly models both absolute utility and relative equity among agents (vehicles). This utility serves as the core decision criterion for selecting the control authority at each time step.

\subsubsection{Fairness Metrics}

We define several fairness metrics to capture diverse aspects of vehicular transit:

\paragraph{Waiting Time Factor}
The waiting time factor $w_i(t)$ for vehicle $i$ at time $t$ is:
\begin{equation}
w_i(t) = \min\biggl(1.0, \frac{\tau_i(t)}{10.0}\biggr),
\end{equation}
where $\tau_i(t)$ denotes the number of consecutive time steps during which vehicle $i$ has been waiting without control authority. This factor prioritizes vehicles that have been waiting longer, capped at 10 time steps to prevent any single vehicle from being overly favored.

\paragraph{Recent Control Ratio}
To avoid monopolization of control authority, we define the recent control ratio $r_i(t)$ as:
\begin{equation}
r_i(t) = \frac{|\{t' \in \mathcal{H}_i : t - t' \leq W\}|}{W},
\end{equation}
where $\mathcal{H}_i$ is the set of time steps in which vehicle $i$ held control authority, and $W$ is the fairness window (typically set to 50 time steps, equivalent to 2.5 seconds with $\Delta t = 0.05\text{ s}$). Smaller $r_i(t)$ values indicate that the vehicle has received less recent control, making it more likely to be assigned authority.

\paragraph{Time-Since-Last-Control}
The urgency factor $u_i(t)$ captures how long it has been since a vehicle last had control:
\begin{equation}
u_i(t) = \min\biggl(1.0, \frac{t - t^{last}_i}{10.0}\biggr),
\end{equation}
where $t^{last}_i$ is the last time step at which vehicle $i$ was granted control authority. Consequently, the longer it has been since a vehicle last exercised control, the longer it must wait before being granted authority again.



\subsubsection{Payoff Function}

We combine these fairness metrics into a payoff function $p_i(t)$ for vehicle $i$ at time $t$:
\begin{equation}
p_i(t) = \alpha_1 r_i(t) + \alpha_2 w_i(t) + \alpha_3 u_i(t),
\end{equation}
where $\alpha_1 + \alpha_2 + \alpha_3 = 1$. This function balances control history, waiting time, urgency, while also accounting for stopping efficiency and velocity uniformity.

\subsubsection{Inequity Aversion Utility (IAU)}

To further enhance fairness, we draw on behavioral economics by incorporating an IAU. This approach recognizes that fairness perception depends both on self-utility and performance relative to others. The IAU for vehicle $i$ at time $t$ is:
\begin{align}
\scriptsize
\text{IAU}_i(t) =\;& p_i(t) 
- \frac{\beta_1}{N-1} \sum_{j \neq i} \max(p_j(t) - p_i(t), 0) \notag \\
& - \frac{\beta_2}{N-1} \sum_{j \neq i} \max(p_i(t) - p_j(t), 0) + \delta \, v_i(t),
\end{align}
where:
\begin{itemize}
    \item $p_i(t)$ is the direct payoff from control authority.
    \item The second term represents disadvantageous inequality, with $\beta_1 > 0$ (typically 1.5) capturing aversion to others having higher payoffs.
    \item The third term reflects advantageous inequality, with $\beta_2 > 0$ (typically 0.5) capturing discomfort when one’s own payoff exceeds others. In general, $\beta_1 > \beta_2$ to reflect greater sensitivity to disadvantageous inequality.
    \item The fourth term adds a bonus proportional to velocity deviation, with $\delta = 0.3$, prioritizing vehicles displaying greater deviation from the mean.
\end{itemize}

\subsubsection{Control Authority Allocation Algorithm}

Algorithm~\ref{alg:fairness_allocation} outlines our fairness-driven control authority allocation procedure. At each time step, the algorithm computes the payoff and IAU values for all eligible vehicles, grants control authority to the vehicle with the highest IAU value, and updates the relevant vehicle states accordingly.

\begin{algorithm}[htbp]
\small
\caption{Fairness-Driven Control Authority Allocation}
\label{alg:fairness_allocation}
\begin{algorithmic}[1]
\REQUIRE Set of vehicles $\mathcal{V}$, total time steps $T$, fairness window $W$
\ENSURE Control authority assignments $\{a_t\}_{t=0}^{T-1}$

\STATE Initialize parameters and historical records
\STATE Assign initial control randomly: $a_0 \leftarrow $ random vehicle from $\mathcal{V}$

\FOR{$t \in \{1, \ldots, T-1\}$}
  \STATE Compute $p_i(t)$ for all $i \in \mathcal{V}$
  
  \STATE Determine eligible set: $\mathcal{E}_t \leftarrow \{i \in \mathcal{V} : r_i(t) < 0.5\}$
  \IF{$\mathcal{E}_t = \emptyset$}
    \STATE $\mathcal{E}_t \leftarrow \mathcal{V}$ \COMMENT{All vehicles if none are below threshold}
  \ENDIF
  
  \STATE Compute $\text{IAU}_i(t)$ for all $i \in \mathcal{E}_t$
  \STATE $a_t \leftarrow \underset{i \in \mathcal{E}_t}{\arg\max} \, \text{IAU}_i(t)$
\ENDFOR

\RETURN $\{a_t\}_{t=0}^{T-1}$
\end{algorithmic}
\end{algorithm}

This method can be seen as a simplified adaptation of the FGT framework \cite{9458933} for intersection management. While the general FGT framework relies on iterative best-response strategies to converge to a Nash equilibrium, our binary decision space (whether a vehicle is granted control or not) allows a single-step solution by choosing the vehicle with the highest IAU. This retains the fairness attributes of FGT while significantly reducing computational overhead. Moreover, by frequently switching the control authority among vehicles, the system ensures that all agents can proceed smoothly through the intersection over time.

\section{Bottom Layer: Safe Real-Time Tracking with Optimal Trajectory Planning via DDP and Safety Filters} \label{sec:bottomlayer}

The bottom layer of our hierarchical control architecture handles real-time trajectory tracking and short-horizon reactive decision-making. It converts the global reference trajectory into feasible low-level commands, ensuring collision avoidance and safety under dynamic interactions.

\subsection{Trajectory Tracking Controllers}
Our trajectory tracking follows the decoupled lateral-longitudinal control structure from the Baidu Apollo platform~\cite{baidu_apollo}, using a discrete-time LQR for lateral and a proportional-derivative (PD) controller for longitudinal control. We also integrate obstacle avoidance via repulsive fields and a CBF-based safety filter.

\subsubsection{Lateral Control via Discrete-Time LQR}

The vehicle's lateral error dynamics are linearized about the reference trajectory, resulting in a discrete-time linear system:
\begin{equation}
x_{k+1} = A_k x_k + B_k \delta_k,
\end{equation}
where \( x_k = [e_{\text{cg}}, \dot{e}_{\text{cg}}, \theta_e, \dot{\theta}_e]^\top \) represents the lateral deviation and heading error along with their derivatives, and \( \delta_k \) is the front steering angle.

To design the discrete-time LQR controller, we first linearize the continuous-time bicycle model along the reference trajectory. The resulting system matrices \( A_k \) and \( B_k \) are then obtained by discretizing the continuous-time linear model. Specifically, the state transition matrix \( A_k \) is computed via Tustin's method, while the input matrix \( B_k \) is obtained using Euler's method for simplicity. The full derivation of the continuous-time matrices \( A_c \) and \( B_c \), including vehicle geometry and tire dynamics considerations, is provided in Appendix~\ref{appendix:linearization}.

We define a quadratic cost function:
\begin{equation}
J = \sum_{k=0}^{\infty} x_k^\top Q x_k + \delta_k^\top R \delta_k,
\end{equation}
and solve the discrete-time Riccati equation for the time-varying system iteratively. Starting with an initial guess \( P_k^{(0)} = Q \), the Riccati recursion is given by:
\begin{equation}
\footnotesize
P_k^{(i+1)} = Q + A_k^\top P_k^{(i)} A_k - A_k^\top P_k^{(i)} B_k (R + B_k^\top P_k^{(i)} B_k)^{-1} B_k^\top P_k^{(i)} A_k,
\end{equation}

Once the sequence \( \{P_k^{(i)}\} \) converges to \( P_k^{*} \) (or reaches a predefined maximum number of iterations), the optimal feedback gain can be computed as:
\begin{equation}
K_k = (R + B_k^\top P_k^{*} B_k)^{-1} B_k^\top P_k^{*} A_k.
\end{equation}

The final control input is then:
\begin{equation}
\delta_k = -K_k x_k + \delta_{\text{ff}},
\end{equation}
where the feedforward term $\delta_{\text{ff}}$ compensates for curvature- and speed-dependent effects:
\begin{equation}
\delta_{\text{ff}} = L \cdot \kappa_{\text{ref}}(k) + \eta(v_k, \kappa_{\text{ref}}),
\end{equation}
with $L$ being the wheelbase and $\eta(\cdot)$ correcting for velocity-dependent effects. Here, $\kappa_{\text{ref}}(k)$ denotes the reference curvature at time step $k$ along the planned path. The feedforward term \(\delta_{\text{ff}}\) reduces steady-state tracking errors, especially at higher speeds. It augments the geometric curvature-based term with additional compensation derived from the vehicle dynamics. A detailed formulation and physical interpretation are provided in Appendix~\ref{appendix:feedforward}.


\subsubsection{Longitudinal Control via PD Regulation}

The longitudinal controller tracks the reference speed $v_{\text{ref}}(k)$ using:
\begin{equation}
a_k = k_p (v_{\text{ref}}(k) - v_k),
\end{equation}
where $a_k$ is the acceleration command and $k_p$ is a proportional gain. In near-stop conditions, the control is overriden by:
\begin{equation}
a_k =
\begin{cases}
a_{\text{strong}} & \text{if } d_k < d_{\text{th}} \text{ and } v_k > v_{\text{th}}, \\
a_{\text{gentle}} & \text{if } d_k < d_{\text{th}} \text{ and } v_k < -v_{\text{th}}, \\
k_p (v_{\text{ref}} - v_k) & \text{otherwise},
\end{cases}
\end{equation}
where \( d_k \) is the distance to the goal, and \( d_{\text{th}}, v_{\text{th}}, a_{\text{strong}}, a_{\text{gentle}} \) are thresholds and acceleration values for near-stop handling.

\subsection{Dynamic Obstacle Avoidance via Repulsive Fields}

To avoid dynamic obstacles, we implement a time-weighted repulsive potential field method that modifies the nominal steering and speed commands. Given the current vehicle position $p_v$ and velocity $v_v$, the predicted position over a finite horizon $\mathcal{T}$ is $p_v(\tau) = p_v + \tau v_v$. Similarly, each dynamic obstacle $o \in \mathcal{O}_t$ is predicted as $p_o(\tau) = p_o + \tau v_o$. For each prediction step $\tau \in \mathcal{T}$, we compute the relative displacement $\Delta p = p_v(\tau) - p_o(\tau)$ and its magnitude $d = \|\Delta p\|$. If $d$ is within the detection range $d_{\text{detect}}$, a repulsive force is generated with time-decaying weight $w(\tau) = \frac{H - \tau}{H}$. The total repulsive force is:

\begin{align}
\vec{F}_{\text{rep}} = 
\begin{bmatrix}
F_x \\
F_y
\end{bmatrix}
= \sum_{o \in \mathcal{O}_t} \sum_{\tau \in \mathcal{T}} w(\tau) 
\left( \frac{1}{d} - \frac{1}{d_{\text{detect}}} \right) 
\frac{\Delta p}{d}.
\end{align}

The vector $\vec{F}_{\text{rep}} = [F_x, F_y]^\top$ indicates the avoidance direction. The avoidance heading deviation is $\Delta \theta_{\text{avoid}} = \text{atan2}(F_y, F_x) - \theta$, where $\theta$ is the vehicle heading. A blending factor $\lambda = \min(1.0, \|\vec{F}_{\text{rep}}\|)$ is used to interpolate between the nominal control and the avoidance action. The final control commands are:

\begin{align}
\delta_k^{\text{mod}} &= (1 - \lambda) \delta_k + \lambda \cdot \Delta \theta_{\text{avoid}}, \\
v_k^{\text{mod}} &= (1 - 0.5\lambda) v_k,
\end{align}

where $\delta_k$ and $v_k$ are the original steering and speed commands, and $\delta_k^{\text{mod}}$, $v_k^{\text{mod}}$ are the modified outputs accounting for obstacle avoidance.

\subsection{Safety-Critical Filtering via CBF-Based QP}

\subsubsection{High Order Control Barrier Function (HOCBF)}
CBFs provide a rigorous mathematical framework for enforcing safety constraints in nonlinear control systems. Here, we consider a time-varying safety set:
\begin{equation}
\mathcal{C}(t) = \{x \in \mathbb{R}^n : h(x, t) \geq 0\},
\end{equation}
where the safety function \( h: \mathbb{R}^n \times \mathbb{R} \rightarrow \mathbb{R} \) is assumed to be continuously differentiable.

We first present the formal definitions of CBFs and HOCBFs as follows:

\textbf{Definition 1 (Control Barrier Function):}  
For a nonlinear control system \( \dot{x} = f(x) + g(x)u \), a continuously differentiable function \( h(x, t) \) is a CBF if there exists a class-\( \mathcal{K} \) function \( \alpha(\cdot) \) such that for all \( x \in \mathcal{C}(t) \),
\begin{equation}
L_f h(x, t) + L_g h(x, t) u + \alpha(h(x, t)) \geq 0,
\end{equation}
where \( L_f \), \( L_g \) denote the Lie derivatives of \( h \) along the vector fields \( f \) and \( g \), respectively.

The relative degree of \( h(x,t) \) with respect to the control input \( u \) is the number of derivatives needed for \( u \) to appear explicitly.

When the relative degree \( m \geq 2 \), the classical CBF is no longer directly applicable, as \( u \) does not appear in the first derivative of \( h \). To address this, we use the HOCBF framework:

\textbf{Definition 2 (High-Order Control Barrier Function):}  
A function \( h(x,t) \) is a HOCBF of relative degree \( m \) \cite{9516971}, if there exist recursively defined functions \( \psi_i(x,t) \), such that:
\begin{align}
\psi_0(x,t) &= h(x,t), \\
\psi_i(x,t) &= \dot{\psi}_{i-1}(x,t) + \alpha_i(\psi_{i-1}(x,t)), \quad i \in \{1,\dots,m\},
\end{align}
where \( \alpha_i(\cdot) \) are class-\( \mathcal{K} \) functions. Associated safe sets are:
\begin{equation}
\mathcal{C}_i(t) = \{ x \in \mathbb{R}^n : \psi_{i-1}(x,t) \geq 0 \}, i \in \{1,\dots,m\}.
\end{equation}
The HOCBF constraint is:
\begin{align}
\label{HOCBF}
\psi_m(x,t) = &\; L_f^m h(x,t) + L_g L_f^{m-1} h(x,t) u \nonumber \\
              &\; + S(h(x,t)) + \alpha_m(\psi_{m-1}(x,t)) \geq 0,
\end{align}
where
\begin{equation}
S(h(x,t)) = \sum_{i=1}^{m-1} L_f^i \left(\alpha_{m-i} \circ \psi_{m-i-1} \right)(x,t),
\end{equation}
and \( \circ \) denotes function composition.

\subsubsection{Flexible High-Order Control Barrier Function (F-HOCBF)}
\label{sec:FHOCBF}

Classical HOCBFs extend CBF theory to safety constraints whose \emph{relative degree} with respect to the control input~$u$ is $m\!\ge\!2$. However, most existing HOCBF formulations generally assume \emph{static, rigid} obstacles and therefore fail to capture realistic traffic scenarios in which obstacle size, velocity, and even \emph{acceleration} evolve over time. Building on the \emph{Flexible CBF} (F-CBF) framework proposed in~\cite{shi2024fairness}, which models velocity-dependent deformation in first-order CBFs, we extend this concept to \emph{High-Order Control Barrier Functions (HOCBFs)} by explicitly embedding obstacle \textbf{geometry, velocity, and acceleration} into the safe-set function. This resulting novel \emph{Flexible HOCBF (F-HOCBF)} formulation handles second-order safety constraints under dynamically evolving obstacle envelopes.

\paragraph{State augmentation}
Let
\begin{equation}
  X =
  \bigl[x^{\top},\;
        p_{ob}^{\top},\;
        v_{ob}^{\top},\;
        a_{ob}^{\top}\bigr]^{\top}
  \in\mathbb R^{n+6},
\end{equation}
where $x\in\mathbb R^{n}$ is the ego-vehicle state, and
$p_{ob},\,v_{ob},\,a_{ob}\in\mathbb R^{2}$ are the obstacle’s planar
position, velocity, and acceleration, respectively.

\paragraph{Adaptive safety ellipse}
As in the F-CBF, the obstacle is covered by an ellipse whose
semi-axes \emph{expand} with both speed and acceleration:
\begin{align}
a_{ob}(t) &=
  \frac{L_{ob}}{2}
  + \mu_1\,\sigma\!\bigl(\|v_{ob,\parallel}\|\!-\!v_0\bigr)\,\|v_{ob,\parallel}\| \\
  &+ \nu_1\,\sigma\!\bigl(\|a_{ob,\parallel}\|\!-\!a_0\bigr)\,\|a_{ob,\parallel}\|,
  \label{eq:FHOCBF_a}\\
b_{ob}(t) &=
  \frac{W_{ob}}{2}
  + \mu_2\,\sigma\!\bigl(\|v_{ob,\perp}\|\!-\!v_0\bigr)\,\|v_{ob,\perp}\|\\
  &+ \nu_2\,\sigma\!\bigl(\|a_{ob,\perp}\|\!-\!a_0\bigr)\,\|a_{ob,\perp}\|,
  \label{eq:FHOCBF_b}
\end{align}
where the sigmoid function is $\sigma(z)=(1+\mathrm e^{-kz})^{-1}$.%
\footnote{%
$L_{ob},\,W_{ob}$ are the nominal length/width;
$v_0,a_0$ are activation thresholds;
$\mu_i,\nu_i,k>0$ are tuning parameters.}
The ellipse is oriented along the instantaneous heading
\(
  \phi(t)=\mathrm{atan2}\!\bigl(v_{ob,y},v_{ob,x}\bigr).
\)

\paragraph{Flexible safe-set function}
For any boundary angle $\sigma\!\in\![0,2\pi]$,
\begin{equation}
\footnotesize
  e(\sigma,t)=
  p_{ob}(t)+
  R\!\bigl(\phi(t)\bigr)
  \begin{bmatrix}
    a_{ob}(t)\cos\sigma\\
    b_{ob}(t)\sin\sigma
  \end{bmatrix},
  \quad
  R(\phi)=
  \begin{bmatrix}
    \cos\phi & -\sin\phi\\
    \sin\phi &  \cos\phi
  \end{bmatrix}.
\end{equation}
Let $p(t)\in\mathbb R^{2}$ be the ego-vehicle position extracted from
$x$.  The minimal clearance
\(
  d_{\min}(t)=\min_{\sigma\in[0,2\pi]}\bigl\|p(t)-e(\sigma,t)\bigr\|
\)
defines the \emph{flexible} safe-set function:
\begin{equation}
  h(X,t)=d_{\min}(t)-d_{\mathrm{safe}},
  \qquad
  d_{\mathrm{safe}}>0.
  \label{eq:FHOCBF_h}
\end{equation}
By construction, $h(X,t)\ge0$ indicates that the ego-vehicle lies \emph{outside} the velocity-/acceleration-inflated obstacle envelope.

\paragraph{Reduction to static obstacles}
Setting \(v_{ob}=a_{ob}=0\) and $a_{ob}=L_{ob}/2$, $b_{ob}=W_{ob}/2$, the envelope collapses to the nominal rectangle-circumscribed ellipse and \(h(X,t)\) coincides with the classical distance-based HOCBF in~\cite{9516971}, ensuring backward compatibility.

\paragraph{Computational aspects}
The inner optimization \(\sigma^\star(t)=\arg\min_{\sigma}d(\sigma)\) is one-dimensional convex problem; a $5$–$8$-grid golden-section search suffices in real time.

\subsubsection{Safety-Filter via F-HOCBF}
Building upon this theoretical foundation, we implement a real-time safety filter using a quadratic program (QP) formulation. This filter enforces safety-critical constraints during trajectory tracking, particularly under dynamic and uncertain obstacle conditions.

To guarantee forward invariance of the safe set while respecting real-time constraints, we follow the CBF-ILQR paradigm in \cite{kong2024adaptive}. The lateral LQR and longitudinal PD controllers provide a nominal control input:
\begin{equation}
u_1 = \begin{bmatrix} \delta_k \\ a_k \end{bmatrix},
\end{equation}
which is optimized for trajectory tracking. However, in the presence of dynamic obstacles or tightened actuation limits, the nominal input \( u_1 \) may violate safety constraints. While repulsive field-based obstacle avoidance can steer the vehicle away from predicted collisions, it provides no formal safety guarantees.

To ensure safety, the final control input is:
\begin{equation}
u_k = u_1 + u_2,
\end{equation}
where \( u_2 \in \mathbb{R}^2 \) is a corrective control computed by solving a CBF-based QP.

Following the structure in \cite{kong2024adaptive}, \cite{9516971} and the safety function $h$ from \eqref{eq:FHOCBF_h}, the corresponding F-HOCBF constraint is:
\begin{equation}
\ddot{h}(x_k) + \alpha_1 \dot{h}(x_k) + \alpha_2 h(x_k) \geq 0,
\end{equation}
which is affine in the total control input \( u = u_1 + u_2 \).

To ensure real-time feasibility and minimal deviation from the nominal control, we solve the following QP at each time step:
\begin{align}
u_2^*, \beta^* = \arg\min_{u_2, \beta} \;\; & \|u_2\|^2 + Q (\beta - \beta_0)^2 \\
\text{s.t.} \quad & A(x_k)(u_1 + u_2) + b(x_k) + \beta \geq 0, \notag\\
& u_{\min} \leq u_1 + u_2 \leq u_{\max}, \quad \beta \geq 0,\notag
\end{align}
where \( \beta \) is a relaxation variable ensuring feasibility, and \(A(x_k) = L_g L_f^{m-1} h(x_k, t)\), \(b(x_k) = L_f^m h(x_k, t) + S(h(x_k, t)) + \alpha_m(\psi_{m-1}(x_k, t))\) are defined from the Lie derivatives of the high-order control barrier function \( h(x,t) \).

The final safe control input is:
\begin{equation}
u_k = u_1 + u_2^*.
\end{equation}
with \( u_2^* \) representing the minimal safety-preserving intervention to \( u_1 \). Since the nominal LQR policy already incorporates repulsive field-based avoidance, \( u_2 \) is often zero or small. As a result, the QP remains computationally lightweight. \textbf{This is why repulsive fields are still necessary even in the presence of a safety filter} — they proactively shape the nominal policy to avoid obstacles, thereby reducing the burden on the correction term and ensuring the overall efficiency of the control framework. The convex formulation further guarantees consistent and computationally efficient solver performance, making the approach suitable for real-time deployment.

\subsection{Execution Loop with Online Safety Tracking Algorithm}

The proposed safety-filtered trajectory tracking scheme operates online, ensuring both real-time feasibility and robust performance in dynamic environments. As outlined in Algorithm~\ref{alg:online_safety_tracking}, it enforces safety constraints with minimal interference to the nominal LQR-PD controller. The decoupled yet coordinated design contributes to computational efficiency.

\begin{algorithm}[htbp]
\small
\caption{Online Safety Tracking with LQR-CBF}
\label{alg:online_safety_tracking}
\begin{algorithmic}[1]
\REQUIRE Vehicle state $x_k$, reference trajectory $\mathcal{T}_{\text{ref}}$, obstacles $\mathcal{O}_t$
\ENSURE Safe control $u_k = [\delta_k, a_k]^\top$

\STATE Compute tracking errors $(e_{\text{cg}}, \theta_e)$ from $\mathcal{T}_{\text{ref}}$
\STATE Linearize dynamics; obtain $A_k$, $B_k$ via discretization
\STATE Solve Riccati equation for gain $K$
\STATE Compute nominal lateral control: $\delta_k \leftarrow -K x_k + \delta_{\text{ff}}$
\STATE Compute longitudinal control: $a_k \leftarrow k_p(v_{\text{ref}} - v_k)$
\STATE Form nominal input $u_1 \leftarrow [\delta_k, a_k]^\top$
\STATE Apply dynamic obstacle avoidance: $u_1^{\text{mod}} \leftarrow \text{ObstacleAvoidance}(u_1, x_k, \mathcal{O}_t)$
\STATE Solve the safety QP (CBF) for minimal correction $u_2^*$
\STATE Compute final safe control: $u_k \leftarrow u_1^{\text{mod}} + u_2^*$
\STATE Update vehicle state: $x_{k+1} \leftarrow f(x_k, u_k)$

\RETURN $u_k$
\end{algorithmic}
\end{algorithm}

\subsection{Differential Dynamic Programming for Optimal Trajectory Planning}

DDP solves nonlinear optimal control problems by iteratively refining a nominal trajectory. At each iteration, DDP linearizes the system dynamics and quadratically approximates the cost function around the current trajectory. The \textbf{backward pass} computes local control updates and feedback gains using dynamic programming, while the \textbf{forward rollout} simulates the system under these updates to generate a new trajectory.

\subsubsection{Trajectory Optimization Objective}

We formulate the trajectory optimization problem using a five-dimensional bicycle model, where the state vector is \( x = [p_x, p_y, \theta, v, \omega]^\top \), representing the vehicle’s position, heading angle, linear velocity, and angular velocity. The control input is \( u = [a, \alpha]^\top \), corresponding to longitudinal and angular acceleration. Given a discrete time step \( h \), the system evolves as \( x_{k+1} = [p_x + h v \cos\theta,\; p_y + h v \sin\theta,\; \theta + h \omega,\; v + h a,\; \omega + h \alpha]^\top \).

The optimization objective is to minimize the cost over a planning horizon of \( N \) steps:
\[
J = \sum_{k=0}^{N-1} \frac{1}{2}(x_k^\top Q x_k + u_k^\top R u_k)h + \frac{1}{2} x_N^\top Q_f x_N,
\]
where \( Q \), \( R \), and \( Q_f \) are positive semi-definite weight matrices penalizing state deviations, control efforts, and terminal state, respectively. To promote safety, additional soft penalties are incorporated to discourage proximity to obstacles, shaping the cost landscape accordingly.

\subsubsection{DDP Iteration: Backward Pass and Forward Rollout}

\paragraph{Backward Pass}
In each iteration, the backward pass computes the derivatives of the value function by propagating them in time using dynamic programming. Linearizing the dynamics along the nominal trajectory yields the matrices $A = \frac{\partial f}{\partial x}$ and $B = \frac{\partial f}{\partial u}$. At time step $k$, the derivatives of the Q-function are computed as:
\begin{align}
Q_x &= \ell_x + A^\top V_x', \quad Q_u = \ell_u + B^\top V_x', \\
Q_{xx} &= \ell_{xx} + A^\top V_{xx}' A, \\
Q_{uu} &= \ell_{uu} + B^\top V_{xx}' B, \\
Q_{ux} &= \ell_{ux} + B^\top V_{xx}' A,
\end{align}
where $\ell(\cdot)$ and its derivatives denote the stage cost, and $V_x'$, $V_{xx}'$ are the gradients of the value function at the subsequent time step.

The optimal perturbation to the control inputs at step $k$ is then:
\begin{equation}
\delta u_k = -Q_{uu}^{-1} Q_u - Q_{uu}^{-1} Q_{ux} \delta x_k.
\end{equation}
To ensure numerical stability and positive definiteness of $Q_{uu}$, we adaptively apply a Levenberg–Marquardt regularization term $\mu I$.

\paragraph{Forward Rollout and Line Search}
Once the control adjustments are obtained, we perform a forward rollout to generate and evaluate the updated state trajectory:
\[
u_k^{\text{new}} = u_k + a \delta u_k,\]
where $a$ is the step size determined through a backtracking line search. The forward simulation computes the actual cost improvement, which is compared with the quadratic model's prediction. The step size $a$ is then adjusted to ensure robust convergence and cost reduction.

This backward-forward iteration continues until either the cost improvement falls below a preset convergence threshold or the step size becomes sufficiently small, at which point a locally optimal trajectory is obtained.





\begin{algorithm}[h]
\small
\caption{Offline Trajectory Optimization with DDP}
\label{alg:ddp_trajectory_planning}
\begin{algorithmic}[1]
\REQUIRE Initial state $x_0$, initial control sequence $\{u_k\}_{k=0}^{N-1}$, cost function $\ell(x,u), \ell_f(x)$, dynamics $f(x,u)$
\ENSURE Optimized state $\{x_k^*\}_{k=0}^N$ and control sequence $\{u_k^*\}_{k=0}^{N-1}$

\STATE Initialize nominal trajectories $\{x_k, u_k\}$ via forward rollout
\REPEAT
    \STATE \textbf{Backward Pass:}
    \FOR{$k = N-1$ to $0$}
        \STATE Linearize dynamics at $(x_k, u_k)$ to get $A_k$, $B_k$
        \STATE Compute derivatives: $Q_x$, $Q_u$, $Q_{xx}$, $Q_{uu}$, $Q_{ux}$
        \STATE Update value function approximation
    \ENDFOR

    \STATE \textbf{Forward Rollout and Line Search:}
    \STATE Set step size $a \leftarrow 1.0$
    \REPEAT
        \STATE Generate new trajectory $\{x_k^{\text{new}}, u_k^{\text{new}}\}$ using $k_k$, $K_k$
        \STATE Evaluate actual and predicted cost improvement
        \STATE Adjust step size $a$ if needed
    \UNTIL{Cost reduction is sufficient or step size too small}

    \STATE Update nominal trajectory with $\{x_k^{\text{new}}, u_k^{\text{new}}\}$
\UNTIL{Convergence criterion is met}

\RETURN $\{u_k^*\}, \{x_k^*\}$
\end{algorithmic}
\end{algorithm}


\section{Experimental Validation and Discussion} \label{sec:experiments}

We validate the fairness-oriented control framework through comparisons with all-way stop and pre-timed signal control.
\vspace{-0.2cm}

\subsection{Experiment Setup}
\begin{figure}[htbp]
\captionsetup{font={footnotesize}}
\captionsetup[subfloat]{labelformat=empty,font=scriptsize} 
\centering
\subfloat[The proposed method]{
    \includegraphics[width=0.5\linewidth]{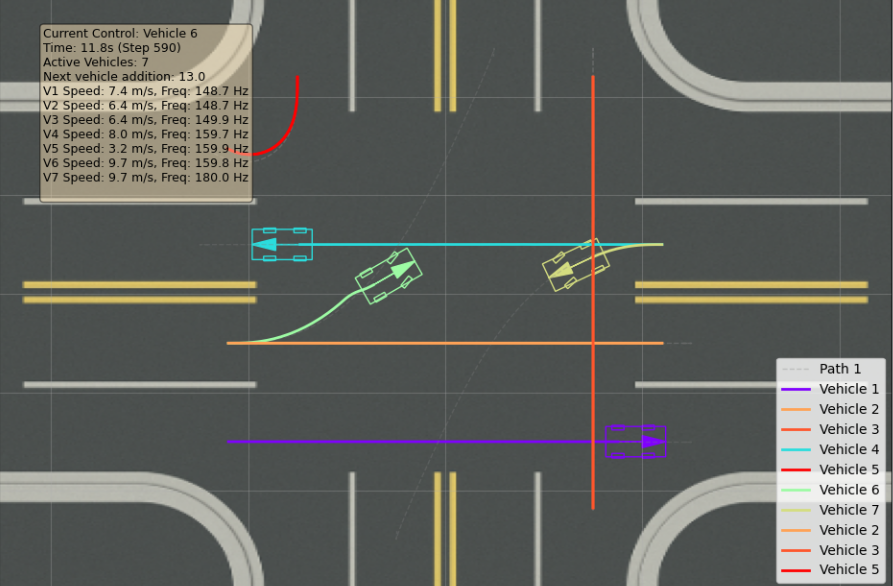}}

\caption{Intersection simulations under medium-demand scenarios and highly unbalanced conditions, with the proposed fairness-aware framework.}
\label{fig:intersection_all_scenarios}
\vspace{-0.6cm}
\end{figure}

To evaluate the performance of the proposed fairness-oriented hierarchical control framework, experiments are conducted in a simulated four-way intersection with two lanes per approach, supporting straight, left-turn, and right-turn maneuvers; see Figure \ref{fig:intersection_all_scenarios}. The simulation platform is implemented in Python and includes vehicle dynamics modeling, motion planning via DDP, fairness-oriented control authority allocation, and real-time collision avoidance via HOCBFs.

Traffic demand is categorized into three levels-low, medium, and high-to represent varying traffic intensities. In addition, three demand distribution patterns across different approaches are considered: balanced, unbalanced, and highly unbalanced, to examine the framework's ability to maintain fairness under heterogeneous inflow conditions. Vehicles enter the simulation at intervals predefined by the demand settings. 

Each vehicle is initialized with its position, velocity, and intended movement (straight, left-turn, or right-turn). The vehicle dynamics follow a bicycle model, with key simulation parameters—including physical properties, control limits, and safety filter coefficients—summarized in Table \ref{parameters}.

\begin{table}[htbp]
\begingroup
\fontsize{6.0pt}{8pt}\selectfont
\centering
\caption{Detailed Simulation Parameters}
\begin{tabular}{l c l c}
\hline
\textbf{Parameter} & \textbf{Value} & \textbf{Parameter} & \textbf{Value} \\
\hline
\multicolumn{4}{l}{\textbf{Simulation Settings}} \\
Sampling Time, $T_s$ & 0.02 s & Max Iterations (LQR) & 150 \\
Convergence Threshold & 0.01 & Fairness Window & 50 timesteps \\
Lane Width & 3.5 m & & \\
\multicolumn{4}{l}{\textbf{Vehicle Physical Parameters}} \\
Vehicle Wheelbase & 2.54 m & Vehicle Width & 1.74 m \\
Vehicle Length & 4.42 m & Front Cornering Stiffness & 155,495 N/rad \\
Rear Cornering Stiffness & 155,495 N/rad & Yaw Inertia & 3,436.24 kg$\cdot$m$^2$ \\
$l_f$ (CG to Front Axle) & 1.165 m & $l_r$ (CG to Rear Axle) & 1.165 m \\
Vehicle Mass & 1,140 kg & & \\
\multicolumn{4}{l}{\textbf{Vehicle Control Parameters}} \\
Max Acceleration & 15.0 m/s$^2$ & Max Steering Angle & 35$^\circ$ (0.611 rad) \\
Max Speed & 65 km/h (18.05 m/s) & LQR $Q$ Matrix & [0.5, 0.3, 1.0, 0.0] \\
LQR $R$ Matrix & [0.75] & Smoothness Weight & 100.0 \\
\multicolumn{4}{l}{\textbf{Safety Filter (HOCBF) Parameters}} \\
Obstacle Radius & 6.0 m & Min Safety Distance & 2.0 m \\
Repulsive Gain & 50.0 & Prediction Horizon & 1.0 s \\
CBF $\alpha_1$ & 2.0 & CBF $\alpha_2$ & 4.0 \\
QP Matrix $P$ & diag(1.0, 1.0) & QP Cost $Q$ & 100 \\
Aux Variable $\beta_0$ & 1.0 & Max Filtered Steering & 35$^\circ$ (0.6109 rad) \\
Max Filtered Accel. & $\pm$15.0 m/s$^2$ & & \\
\hline
\end{tabular}
\label{parameters}
\vspace{-0.2cm}
\endgroup
\end{table}

Figure \ref{fig:intersection_all_scenarios} illustrates the simulated intersection environment under high-demand scenarios. During each simulation run, vehicles share their states with the centralized fairness-oriented control allocator, which dynamically assigns control authority based on fairness metrics including waiting times, recent control ratios, urgency, and velocity deviations. The bottom-layer controller employs LQR and CBFs to ensure real-time, collision-free trajectory tracking.

\subsection{Evaluation Metrics}

To comprehensively evaluate the proposed hierarchical framework, we assess performance across four categories: fairness, safety, efficiency, and real-time performance. Each category includes quantitative metrics reflecting system performance in realistic intersections.

\subsubsection{Fairness Metrics}

\textbf{Jain's Fairness Index (JFI):}  
Evaluates the equality of control authority distribution among all participating vehicles:
\begin{equation}
\text{JFI}(t) = \frac{\left(\sum_{i=1}^{N} c_i(t)\right)^2}{N \cdot \sum_{i=1}^{N} \left(c_i(t)\right)^2},
\end{equation}
where $c_i(t)$ is the cumulative number of times vehicle $i$ has been granted control up to time $t$. A value of 1 indicates perfect fairness, while values closer to $1/N$ signify greater inequality.

\textbf{Gini Coefficient:}  
Quantifies inequality in control authority distribution:
\begin{equation}
\text{Gini}(t) = \frac{1}{N}\left(N + 1 - 2 \cdot \frac{\sum_{i=1}^{N} (N+1-i)c_i^*(t)}{\sum_{i=1}^{N} c_i^*(t)}\right),
\end{equation}
where $c_i^*(t)$ are the control counts sorted in ascending order. Lower values indicate more equitable distributions; A value of 0 indicates perfect fairness. \footnote{In our setup, offline planned trajectories are normalized to have identical nominal travel times and constant speeds. Therefore, without interactions or disturbances, all vehicles would accrue control authority at the same rate; that is, $c_i(t)=c_j(t)$ for all $i,j$, and $c_i^*(t)=c_j^*(t)$. Otherwise, the JFI and Gini coefficient would not serve as valid fairness measures.} While both measure fairness, JFI emphasizes overall equality across agents, whereas the Gini coefficient is more sensitive to individual disparities at the extremes.

\subsubsection{Safety Metrics}

\textbf{Minimum Inter-Vehicle Distance:}  
The smallest observed distance between any two vehicles during the simulation. 
\textbf{Average Minimum Distance:}  
The mean of the smallest pairwise distances observed across all time steps and vehicle pairs, indicating overall spacing quality.
\textbf{Critical Distance Events:}  
The number of time steps where inter-vehicle distance falls below a predefined critical threshold (e.g., 2 $m$), signaling near-miss events.
\textbf{Total Collisions:}  
It represent the total number of collision events that occurred during the simulation.

\subsubsection{Efficiency Metrics}
\textbf{Throughput:}  
The total number of vehicles that complete their intended paths through the intersection during the simulation.
\textbf{Average Delay:}  
The average waiting time for each vehicle to cross the intersection, relative to free-flow travel.
\textbf{Maximum and Minimum Delay:}  
The longest and shortest observed delays, indicating the worst- and best-case transit times.
\textbf{Delay Standard Deviation:}  
Variation in delay among vehicles; lower value implies more uniform waiting times.

\subsubsection{Real-Time Performance Metrics}
\textbf{Controller Frequency:}  
Measured in~Hz, it includes the average, maximum, minimum, and 10--90\% range of control loop execution rate, characterizing the system's responsiveness and consistency.


\subsection{Experiment at Unsignalized Intersection}

The proposed fairness-aware hierarchical intersection control framework is evaluated across three traffic demand levels (low, medium, and high) and demand distributions (balanced, unbalanced, and highly unbalanced), and compared against a baseline all-way stop control.

\subsubsection{Fairness Performance}
\begin{figure}[htbp]
\captionsetup{font={footnotesize}}
\centering
\includegraphics[width=0.95\linewidth]{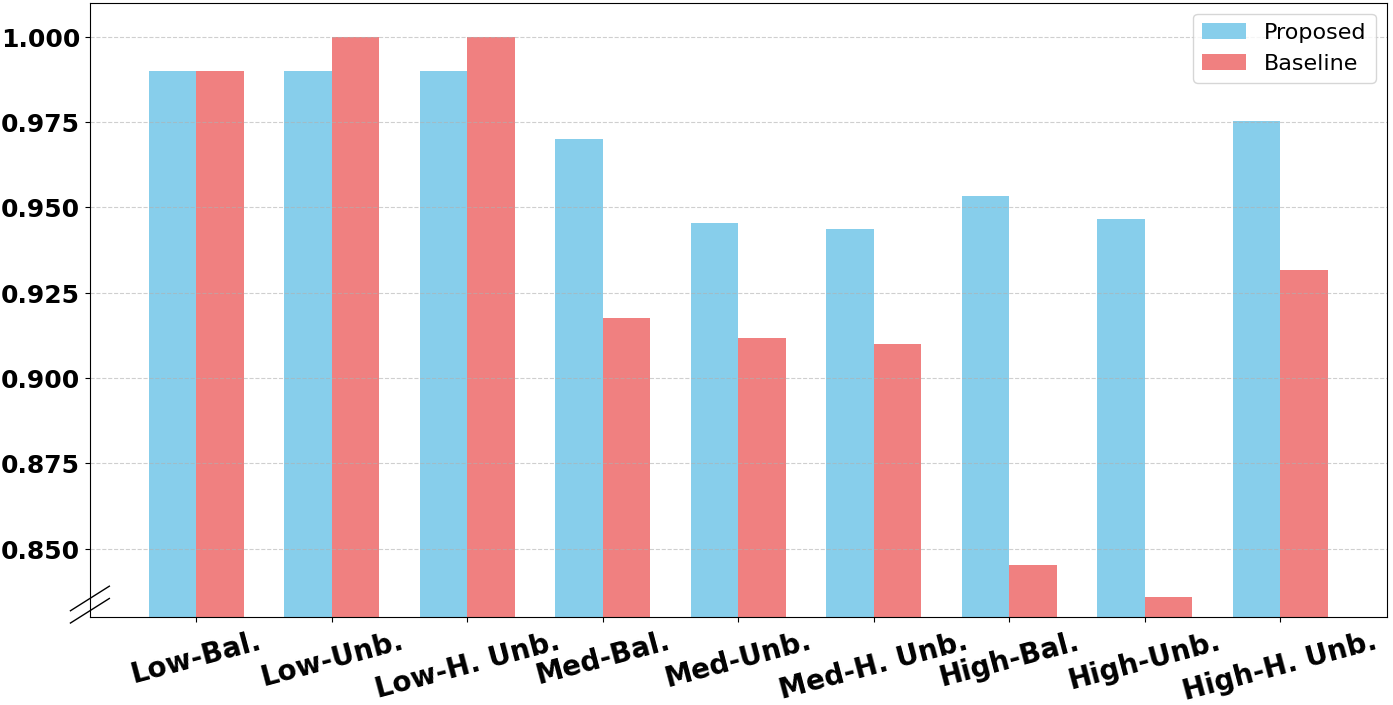}
\vspace{-0.4cm}
\caption{Jain's fairness index across traffic scenarios}
\label{fig:Jains}
\vspace{-0.4cm}
\end{figure}

\begin{figure}[htbp]
\captionsetup{font={footnotesize}}
\centering
\includegraphics[width=0.95\linewidth]{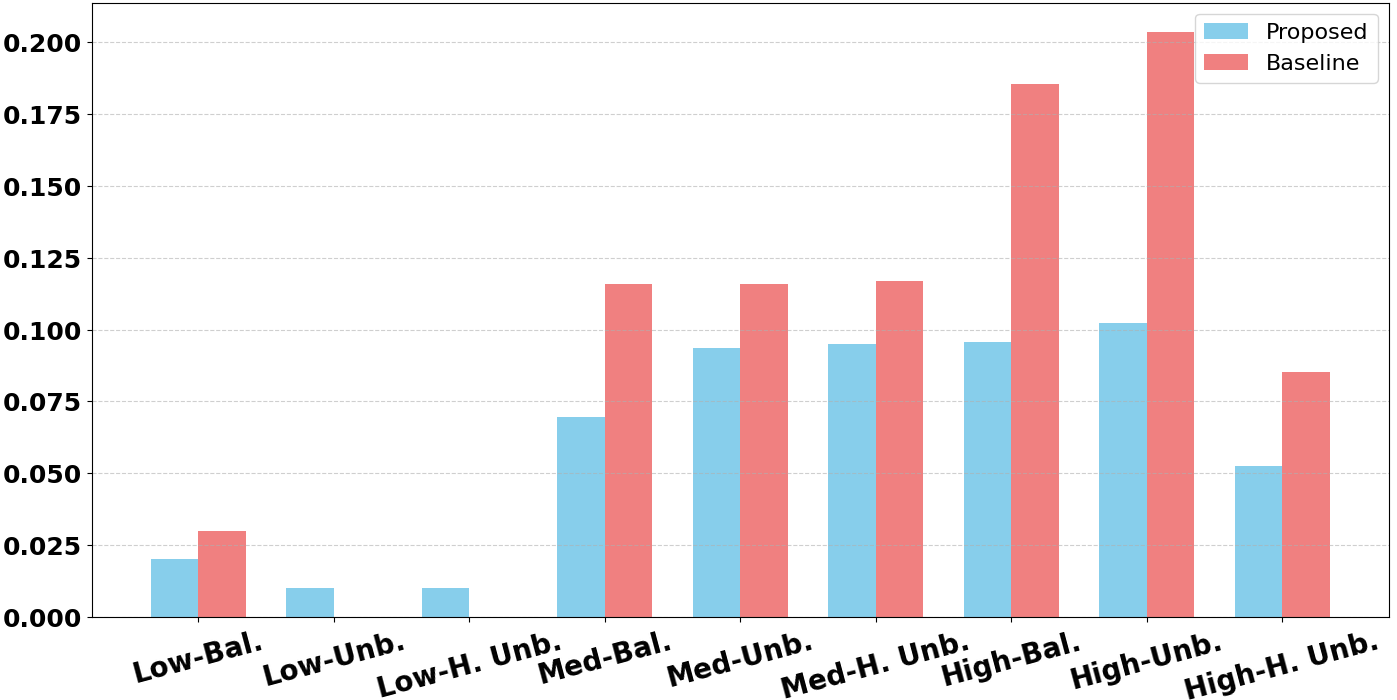}
\vspace{-0.4cm}
\caption{Gini coefficient across traffic scenarios}
\label{fig:Gini}
\vspace{-0.4cm}
\end{figure}

As shown in Figs. \ref{fig:Jains} and \ref{fig:Gini}, under low traffic demand, both methods achieved near-perfect fairness, with JFI values close to~1.0 and Gini coefficients near zero, indicating an almost equal distribution of crossing opportunities. As traffic demand increased, fairness disparities became more pronounced. 
Under medium demand, the proposed method achieved higher JFI values 
($\sim$0.95 vs.~0.91) and lower Gini coefficients (0.08 vs.~0.12), indicating better equity. 
At high demand, the gap widened: the proposed method maintained JFI around~0.95, 
while the baseline dropped to~0.84, with consistently lower Gini values.

With unbalanced demand distributions, fairness declined for both methods, but the proposed method consistently outperformed the baseline. In the most challenging case—high demand with a highly unbalanced distribution—the proposed method still achieved JFI~$\approx$~0.98 compared to the baseline’s~0.93, and a lower Gini coefficient (0.05 vs.~0.09). Overall, these results confirm that the proposed fairness-aware framework substantially enhances fairness in intersection management, particularly under heavy and unbalanced traffic.

\subsubsection{Safety Performance}
Table \ref{experiment-results} summarizes the safety and efficiency performance. Across all scenarios, the proposed method achieved zero collisions, demonstrating robust collision avoidance. Smaller inter-vehicle distances—especially under medium and high traffic demand—indicate efficient spatial use, though they led to more frequent critical-distance events. Nevertheless, no collisions occurred, confirming safe operation under dense, multi-vehicle interactions.

\subsubsection{Efficiency Performance}
The proposed method substantially improved efficiency, particularly under medium and high traffic demand. In the high-demand, highly unbalanced case, throughput reached 3,480~veh/hr compared to 1,440~veh/hr for the baseline, while under medium-demand, unbalanced conditions it achieved 1,860~veh/hr versus 1,470~veh/hr. Average delays for the proposed method were consistently lower across all scenarios. Overall, the proposed framework achieved higher throughput, shorter and more stable delays, and better scalability under increasing traffic demand and more unbalanced distributions. It efficiently utilizes space to enhance efficiency, while the repulsive fields and safety filter ensure collision-free safety despite more critical events and smaller inter-vehicle gaps.

\begin{table*}[htbp]
\small
\renewcommand{\arraystretch}{1.1}
\centering
\caption{Comprehensive Simulation Results for Fairness-Aware Intersection Control Across Different Traffic Scenarios}
\label{experiment_results}
\small
\resizebox{\textwidth}{!}{%
\begin{tabular}{l|cc|cc|cc|cc|cc|cc|cc|cc|cc}
\toprule
& \multicolumn{6}{c|}{\textbf{Low Traffic Demand (990 veh/hr)}} & \multicolumn{6}{c|}{\textbf{Medium Traffic Demand (2,010 veh/hr)}} & \multicolumn{6}{c}{\textbf{High Traffic Demand (3,600 veh/hr)}} \\
\cmidrule(lr){2-7} \cmidrule(lr){8-13} \cmidrule(lr){14-19}
& \multicolumn{2}{c|}{\textbf{Balanced}} & \multicolumn{2}{c|}{\textbf{Unbalanced}} & \multicolumn{2}{c|}{\textbf{Highly Unbal.}} & \multicolumn{2}{c|}{\textbf{Balanced}} & \multicolumn{2}{c|}{\textbf{Unbalanced}} & \multicolumn{2}{c|}{\textbf{Highly Unbal.}} & \multicolumn{2}{c|}{\textbf{Balanced}} & \multicolumn{2}{c|}{\textbf{Unbalanced}} & \multicolumn{2}{c}{\textbf{Highly Unbal.}} \\
\cmidrule(lr){2-3} \cmidrule(lr){4-5} \cmidrule(lr){6-7}
\cmidrule(lr){8-9} \cmidrule(lr){10-11} \cmidrule(lr){12-13}
\cmidrule(lr){14-15} \cmidrule(lr){16-17} \cmidrule(lr){18-19}
\textbf{Metric} & \textbf{Prop.} & \textbf{Base.} & \textbf{Prop.} & \textbf{Base.} & \textbf{Prop.} & \textbf{Base.} & \textbf{Prop.} & \textbf{Base.} & \textbf{Prop.} & \textbf{Base.} & \textbf{Prop.} & \textbf{Base.} & \textbf{Prop.} & \textbf{Base.} & \textbf{Prop.} & \textbf{Base.} & \textbf{Prop.} & \textbf{Base.} \\
\midrule
\multicolumn{19}{c}{\textbf{Safety Metrics}} \\
\midrule
Min. Inter-Veh. Dist. (m) & 1.58 & 3.48 & 1.70 & 3.52 & 1.60 & 4.98 & 0.14 & 3.52 & 0.16 & 3.55 & 0.14 & 3.48 & 3.42 & 3.47 & 1.94 & 4.98 & 0.14 & 3.54 \\
Avg. Min. Distance (m) & 11.25 & 15.42 & 12.30 & 12.66 & 12.08 & 10.25 & 4.55 & 9.18 & 4.90 & 9.48 & 2.30 & 8.60 & 4.35 & 7.95 & 4.40 & 8.18 & 4.25 & 9.30 \\
Critical Events (s) & 14 & 0 & 11 & 0 & 0 & 0 & 15 & 0 & 19 & 0 & 40 & 0 & 0 & 0 & 20 & 0 & 35 & 0 \\
Total Collisions & 0 & 0 & 0 & 0 & 0 & 0 & 0 & 0 & 0 & 0 & 0 & 0 & 0 & 0 & 0 & 0 & 0 & 0 \\
\midrule
\multicolumn{19}{c}{\textbf{Efficiency Metrics}} \\
\midrule
Throughput (veh/hr) & 960 & 900 & 960 & 930 & 960 & 930 & 1980 & 1770 & 1860 & 1470 & 1800 & 1410 & 3540 & 1800  &  3360   & 1470 &  3480   & 1440 \\
Avg. Delay (s) & 0.24 & 0.50 & 0.23 & 0.70 & 0.23 & 0.70 & 4.88 & 6.10 & 4.82 & 6.08 & 4.76 & 5.52 & 2.18 & 5.28 & 3.70 & 4.52 & 3.92 & 5.06 \\
Max. Delay (s) & 0.36 & 0.74 & 0.34 & 1.76 & 0.36 & 1.76 & 9.12 & 9.68 & 9.10 & 9.68 & 9.08 & 9.44 & 3.58 & 9.28 & 5.44 & 9.28 & 5.40 & 8.90 \\
Min. Delay (s) & 0.10 & 0.36 & 0.12 & 0.36 & 0.10 & 0.36 & 2.58 & 3.00 & 2.64 & 3.02 & 2.46 & 2.78 & 0.34 & 3.10 & 1.58 & 1.36 & 1.22 & 2.86 \\
Delay Std. Dev. (s) & 0.13 & 0.18 & 0.13 & 0.59 & 0.12 & 0.59 & 2.62 & 2.84 & 2.64 & 2.83 & 2.62 & 2.73 & 1.26 & 2.66 & 1.26 & 2.55 & 1.28 & 1.90 \\
\midrule
\end{tabular}
} 
\vspace{-3mm}
\footnotesize
\begin{flushleft}
\centering
\textbf{Prop.}: Proposed fairness-aware method; \textbf{Base.}: Baseline all-stop control.\\
Demand balance ratios: Balanced = 1:1:1:1; Unbalanced = 4:2:1:1; Highly Unbalanced = 4:3:1:0.
\end{flushleft}
\label{experiment-results}
\vspace{-0.6cm}
\end{table*}

\subsubsection{Real-Time Computational Feasibility}
\begin{figure}[htbp]
\captionsetup{font={footnotesize}}
\centering
\includegraphics[width=1.0\linewidth]{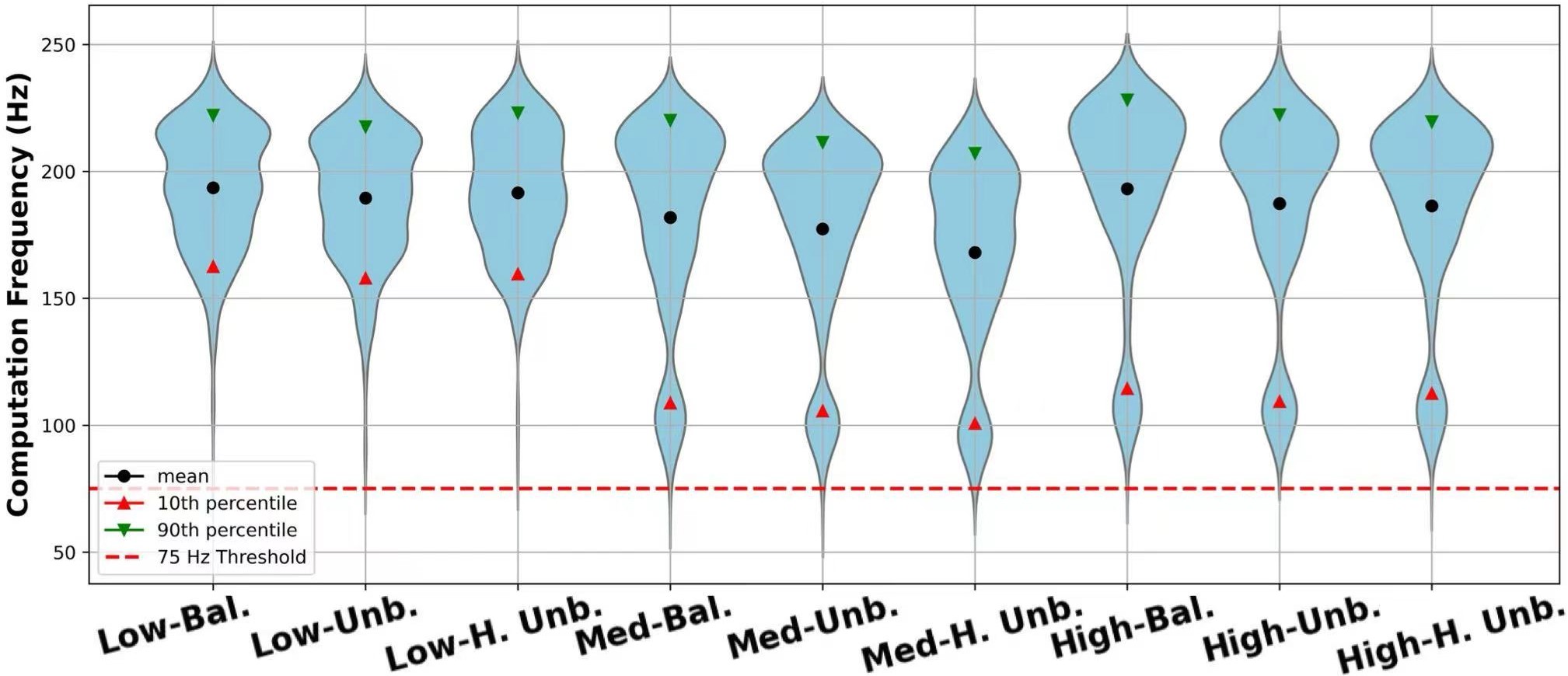}
\vspace{-0.6cm}
\caption{Real-time control frequency distribution}
\label{fig:frequency}
\vspace{-0.6cm}
\end{figure}

The proposed method also demonstrated strong real-time feasibility with average control loop frequencies ranging from 168 Hz to 193 Hz across all scenarios (Fig. \ref{fig:frequency}). Rather than relying on minimum and maximum values—often distorted by transient spikes or initialization artifacts—we evaluated runtime consistency using the 10th to 90th percentile control frequencies. In all scenarios, the 10th percentile frequency exceeded 100 Hz, with most of the distribution remaining above the 75 Hz threshold. These results confirm that the proposed framework maintains high and stable execution rates, ensuring computational efficiency and suitability for real-time deployment.

\subsection{Experiment at Signalized Intersection}





In the signalized intersection experiment, the cycle length and signal phases were configured according to the Manual on Uniform Traffic Control Devices (MUTCD)\cite{traffic2009manual}. The proposed hierarchical framework was compared with the baseline pre-timed signal operation under high traffic demand (3,600 veh/hr), summarized in Table \ref{tab:traffic_light_comparison}. The proposed method consistently achieved higher fairness across all demand distributions (with JFI $> 0.94$, lower Gini). It also improved efficiency, achieving higher throughput and over 40\% shorter average delays. Similar to the all-way-stop experiment, it efficiently uses space to improve efficiency while ensuring safety despite more critical events and smaller gaps.
Meanwhile, the control loop sustained high update rates, ensuring reliable real-time performance; baseline results were measured only during vehicle motion, excluding signal-induced stops. These results confirm that the proposed framework generalizes effectively across both unsignalized and signalized intersection scenarios.


\begin{table}[htbp]
\caption{Signalized Intersection with Varying Demand Distributions}
\centering
\scriptsize
\setlength{\tabcolsep}{3pt}
\renewcommand{\arraystretch}{0.95}
\begin{tabular}{lcccccc}
\toprule
\textbf{Metrics} & \multicolumn{2}{c}{\textbf{Balanced}} & \multicolumn{2}{c}{\textbf{Unbalanced}} & \multicolumn{2}{c}{\textbf{High-Unbalanced}} \\
\cmidrule(lr){2-3} \cmidrule(lr){4-5} \cmidrule(lr){6-7}
& \textbf{Prop.} & \textbf{Base.} & \textbf{Prop.} & \textbf{Base.} & \textbf{Prop.} & \textbf{Base.} \\
\midrule
\multicolumn{7}{c}{\textbf{Fairness Metrics}} \\
\midrule
Jain's Fairness Index     & 0.964 & 0.880 & 0.945 & 0.826 & 0.954 & 0.803 \\
Gini Coefficient          & 0.086 & 0.139 & 0.111 & 0.160 & 0.099 & 0.150 \\
\midrule
\multicolumn{7}{c}{\textbf{Safety Metrics}} \\
\midrule
Min. Inter-Veh. Dist. (m) & 1.48 & 2.74 & 0.56 & 0.88 & 0.27 & 1.02 \\
Avg. Min. Distance (m)    & 4.62 & 5.58 & 1.02 & 2.10 & 1.01 & 3.50 \\
Critical Events (s)   & 5    & 2    & 22   & 13   & 42   & 28 \\
Total Collisions          & 0    & 0    & 0    & 0    & 0    & 0 \\
\midrule
\multicolumn{7}{c}{\textbf{Efficiency Metrics}} \\
\midrule
Throughput (veh/hr)       &  3540   & 3240  &  3360   & 3000 &  3480   & 2400 \\
Avg. Delay (s)        & 2.40 & 5.52 & 3.78 & 6.74 & 4.06 & 7.12 \\
Max. Delay (s)        & 4.00 & 7.08 & 5.60 & 8.60 & 6.24 & 9.18 \\
Min. Delay (s)        & 0.50 & 0.00 & 1.20 & 0.12 & 0.98 & 0.00 \\
Delay Std. Dev. (s)   & 1.30 & 2.64 & 1.44 & 3.10 & 1.56 & 3.18 \\
\midrule
\multicolumn{7}{c}{\textbf{Real-Time Control Metrics}} \\
\midrule
Avg. Ctrl. Freq. (Hz)       & 172.85 & 179.62 & 182.10 & 176.50 & 153.45 & 148.24 \\
Min. Ctrl. Freq. (Hz)       & 101.25 & 95.73  & 64.75  & 68.16  & 55.20  & 51.99  \\
Max. Ctrl. Freq. (Hz)       & 246.95 & 238.77 & 237.40 & 241.19 & 251.80 & 240.22 \\
10th Pctl. Ctrl. Freq. (Hz) & 135.00 & 138.00 & 142.00 & 130.00 & 105.00 & 108.00 \\
90th Pctl. Ctrl. Freq. (Hz) & 225.00 & 222.00 & 215.00 & 220.00 & 195.00 & 190.00 \\
\bottomrule
\end{tabular}
\label{tab:traffic_light_comparison}
\vspace{-0.6cm}
\end{table}

\section{Conclusion} \label{sec:conclusion}
This paper proposed a hierarchical control architecture that integrates \emph{fairness}, \emph{safety}, and \emph{real-time feasibility} for CAVs at intersections. An inequity-aversion allocator selects a single vehicle to receive control authority at each step; while the selected vehicle tracks an offline DDP trajectory with a curvature-aware LQR/PD controller. A F-HOCBF filter guarantees real-time safety under dynamic conditions.  

Comprehensive simulations across low-, medium-, and high-traffic demands, as well as balanced to highly unbalanced demand distributions, validated the framework's effectiveness. The JFI remained above 0.94 and the Gini coefficient below 0.12 in all tests. Intersection throughput increased by up to a factor of two compared with the all-way-stop and signalized baselines, average delay was reduced by as much as 60\%, and no collisions occurred. The control frequency consistently exceeded 50~Hz, demonstrating reliable real-time performance.

These results highlight the potential of the proposed fairness-aware hierarchical control as a viable and deployable solution for intersection management. Its modular and computationally efficient design facilitates practical implementation and future extensions. Future work will explore applications in mixed-autonomy environments, large-scale field experiments, and integration into broader urban mobility systems.




\appendices

\section{Linearization of the Dynamics Model}
\label{appendix:linearization}

The matrices \( A_k \) and \( B_k \) are obtained by discretizing the continuous-time linearized bicycle model. The state-transition matrix \( A_k \) is computed using Tustin's method (bilinear transform):
\begin{equation}
\small
A_k = \left(I - \frac{T_s}{2} A_c\right)^{-1} \left(I + \frac{T_s}{2} A_c\right),
\end{equation}
where \( A_c \) is the continuous-time system matrix and \( T_s \) is the control time step. The input matrix \( B_k \) is discretized using Euler's method for simplicity:
\begin{equation}
B_k = T_s B_c,
\end{equation}
where \( B_c \) is the continuous-time input matrix. For improved consistency with \(A_k\), Tustin's method may also be applied:
\begin{equation}
B_k = T_s (I - \tfrac{T_s}{2} A_c)^{-1} B_c.
\end{equation}

The continuous-time system matrix \( A_c \in \mathbb{R}^{4 \times 4} \) and input matrix \( B_c \in \mathbb{R}^{4 \times 1} \) are derived by linearizing the bicycle model about the reference trajectory, taking into account the vehicle's geometry and tire dynamics. Let \( m = m_f + m_r \) denote the total vehicle mass, and \( I_z \) the yaw moment of inertia. Then \( A_c \) is given by:
\begin{equation}
\small
A_c = 
\begin{bmatrix}
0 & 1 & 0 & 0 \\
0 & -\dfrac{c_f + c_r}{m v} & \dfrac{c_f + c_r}{m} & \dfrac{l_r c_r - l_f c_f}{m v} \\
0 & 0 & 0 & 1 \\
0 & \dfrac{l_r c_r - l_f c_f}{I_z v} & \dfrac{l_f c_f - l_r c_r}{I_z} & -\dfrac{l_f^2 c_f + l_r^2 c_r}{I_z v}
\end{bmatrix}
\end{equation}

\begin{equation}
B_c = 
\begin{bmatrix}
0 & \dfrac{c_f}{m} & 0 & \dfrac{l_f c_f}{I_z}
\end{bmatrix}^{\!\top}
\end{equation}


\section{Derivation of the Curvature-Dependent Feedforward Term}
\label{appendix:feedforward}

The feedforward steering term \(\delta_{\text{ff}}\) compensates for the steady-state lateral tracking error that would remain if the controller relied solely on feedback.  
At low speeds, the kinematic bicycle model indicates that a steering angle of \(L\,\kappa_{\text{ref}}\) (wheelbase times curvature) is sufficient to follow a path of curvature \(\kappa_{\text{ref}}\).  
However, at higher speeds, lateral tire slip causes an under- or over-steer bias that grows approximately with \(v^{2}\kappa_{\text{ref}}\). To address this, the geometric steering angle is augmented with a velocity-dependent correction:
\begin{equation}
\delta_{\text{ff}}
= L\,\kappa_{\text{ref}}
+ k_v\,v^{2}\kappa_{\text{ref}}
- K_{k}\left(
    l_r\,\kappa_{\text{ref}}
    -
    \frac{l_f\,m\,v^{2}\kappa_{\text{ref}}}{2c_r L}
\right)
\end{equation}
where
\begin{equation}
k_v = \frac{l_r m}{2c_f L} - \frac{l_f m}{2c_r L}, \quad
L = l_f + l_r.
\end{equation}

Here, \(L\kappa_{\text{ref}}\) and \(k_v v^2 \kappa_{\text{ref}}\) compensate for curvature- and velocity-induced deviations based on a linear tire model.  
The feedback gain \(K_{k}\) further suppresses coupling between yaw rate and lateral displacement that arises in high-speed, high-curvature maneuvers. Additionally, when the vehicle is reversing (i.e., gear set to \texttt{GEAR\_REVERSE}), the sign of the velocity-related term is inverted to match the reversed kinematic configuration.  
This composite feedforward design improves steady-state accuracy without adding feedback-induced oscillations.

 




\vspace{-0.3cm}  
{\small       
\renewcommand{\baselinestretch}{0.8}\selectfont  
\bibliographystyle{IEEEtran}
\bibliography{ref}
}

\vfill

\end{document}